\documentclass{article}

\usepackage{microtype}
\usepackage{graphicx}
\usepackage{subfigure}
\usepackage{booktabs} 
\usepackage{enumitem}
\usepackage{multirow}
\usepackage{makecell}
\usepackage{xcolor}
\definecolor{color+}{RGB}{0, 100, 0}
\definecolor{color-}{RGB}{200, 0, 0}

\usepackage{hyperref}

\usepackage[accepted]{icml2025}

\usepackage{amsmath}
\usepackage{amssymb}
\usepackage{mathtools}
\usepackage{amsthm}

\usepackage[capitalize,noabbrev]{cleveref}

\usepackage{eso-pic} 

\newcommand{\longcatlogoposfirst}{\AtPageUpperLeft{\hspace{20mm}\raisebox{-25mm}{\includegraphics[height=9mm]{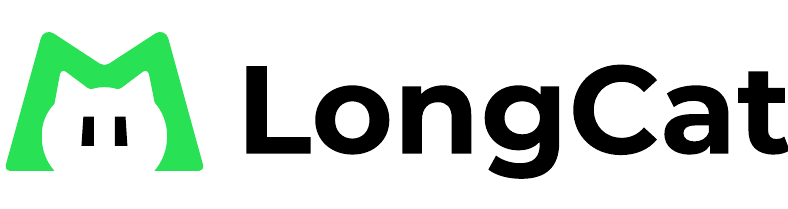}}}}
\newcommand{\longcatlogoposother}{\AtPageUpperLeft{\hspace{168mm}\raisebox{-20mm}{\includegraphics[height=5.8mm]{figures/longcat-logo-full.pdf}}}}
\AddToShipoutPictureFG{%
  \ifnum\value{page}=1\relax
    \longcatlogoposfirst
  \else
    \longcatlogoposother
  \fi
}

\theoremstyle{plain}

\theoremstyle{definition}

\theoremstyle{remark}

\usepackage[textsize=tiny]{todonotes}
\definecolor{mygreen}{RGB}{34,139,34}
\definecolor{myred}{RGB}{178,34,34}

\icmltitlerunning{Learning to Self-Verify Makes Language Models Better Reasoners}

\begin{document}

\twocolumn[
\icmltitle{Learning to Self-Verify Makes Language Models Better Reasoners}




\begin{icmlauthorlist}
\icmlauthor{Yuxin Chen}{NUS}
\icmlauthor{Yu Wang}{USTC}
\icmlauthor{Yi Zhang}{USTC}
\icmlauthor{Ziang Ye}{USTC}
\icmlauthor{Zhengzhou Cai}{BUPT}
\icmlauthor{Yaorui Shi}{USTC} \\
\icmlauthor{Qi Gu}{Meituan}
\icmlauthor{Hui Su}{Meituan}
\icmlauthor{Xunliang Cai}{Meituan}
\icmlauthor{Xiang Wang}{USTC}
\icmlauthor{An Zhang}{USTC}
\icmlauthor{Tat-Seng Chua}{NUS}
\end{icmlauthorlist}

\icmlaffiliation{NUS}{National University of Singapore}
\icmlaffiliation{USTC}{University of Science and Technology of China}
\icmlaffiliation{BUPT}{Beijing University of Posts and Telecommunications}
\icmlaffiliation{Meituan}{Meituan}

\icmlcorrespondingauthor{An Zhang}{an\_zhang@ustc.edu.cn}
\icmlcorrespondingauthor{Qi Gu}{guqi03@meituan.com}

\icmlkeywords{Machine Learning, ICML}

\vskip 0.3in
]



\printAffiliationsAndNotice{} 

\begin{abstract}
Recent large language models (LLMs) achieve strong performance in generating promising reasoning paths for complex tasks.
However, despite powerful generation ability, LLMs remain weak at verifying their own answers, revealing a persistent capability asymmetry between generation and self-verification.
In this work, we conduct an in-depth investigation of this asymmetry throughout training evolution and show that, even on the same task, improving generation does not lead to corresponding improvements in self-verification.
Interestingly, we find that the reverse direction of this asymmetry behaves differently: learning to self-verify can effectively improve generation performance, achieving accuracy comparable to standard generation training while yielding more efficient and effective reasoning traces.
Building on this observation, we further explore integrating self-verification into generation training by formulating a multi-task reinforcement learning framework, where generation and self-verification are optimized as two independent but complementary objectives.
Extensive experiments across benchmarks and models demonstrate performance gains over generation-only training in both generation and verification capabilities. 
Our code is publicly available at \url{https://github.com/chenyuxin1999/Learning-to-Self-Verify}.

\end{abstract}

\section{Introduction}
\label{sec:introduction}

Large language models (LLMs) have demonstrated strong capabilities in complex reasoning~\cite{Deepseek-R1, Qwen3, Gemma3, GPT-5.2}.
With the advancement of Reinforcement Learning with Verifiable Rewards (RLVR), current models have made substantial progress on verifiable tasks such as mathematics and programming~\cite{deepseek-math-v2, Claude-Opus-4.5, GLM-4.7}, while also showing consistent improvements on open-domain tasks including writing, dialogue, and general problem solving~\cite{DeepSeek-V3.2, MiniMax-M2, open_domain1, open_domain2}.
Despite these advances, a fundamental asymmetry remains: even the most powerful models often lack the ability to reliably verify the correctness of their own outputs.

\begin{figure}[t]
    \centering
    \includegraphics[width=0.48\textwidth]{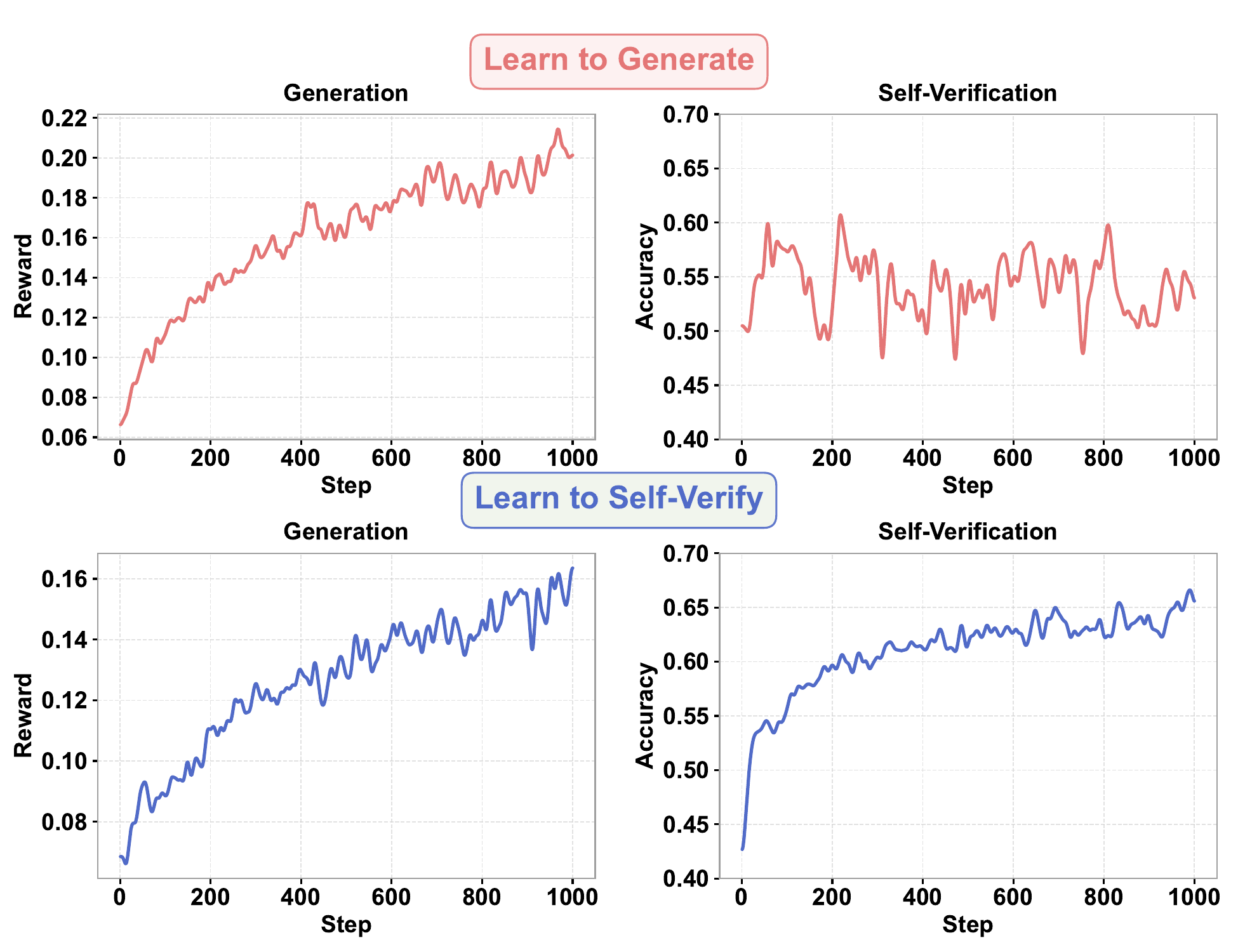}
\vspace{-1.5em}
\caption{
Training dynamics of Qwen2.5-1.5B-Instruct.
(\textbf{Top}) It reveals a persistent asymmetry between generation and self-verification: learning to generate does not lead to improved self-verification ability, even on the same task.
(\textbf{Down}) In the reverse direction, learning to self-verify not only improves self-verification ability but also leads to improved generation performance.
}
\label{fig:asymmetry_intro}
\end{figure}

LLMs have long been considered incapable of verifying the correctness of their own answers~\cite{cant_self_verify1, cant_self_verify2, cant_self_verify3}. 
With the advent of RLVR, some works observe that models can exhibit emergent self-verification behaviors, sometimes also referred to as an ``aha moment''~\cite{Deepseek-R1, aha_1, aha_2}. 
However, subsequent analyses suggest that most of these behaviors are in fact fake verification: although the model appears to be checking its previous reasoning, this step has little impact on the final answer, fundamentally due to the model’s limited ability to reliably verify its own generations~\cite{fake_verification1, fake_verification2}.
More importantly, self-verification capability does not naturally improve with increased model scale or stronger generation ability~\cite{cant_scale}, revealing a persistent asymmetry between generation and self-verification.
Motivated by this, several approaches attempt to jointly optimize generation and verification within the same training step, treating verification as an auxiliary component~\cite{join_train1, join_train2, join_train3}.
In practice, however, the training dynamics of these methods remain dominated by the generation objective, leaving the fundamental asymmetry largely unexplored.

In this work, we conduct an in-depth investigation of the asymmetry between generation and self-verification.
Specifically, we explicitly train the LLM to generate better answers in a specific domain (e.g., mathematics) and track how it behaves when verifying its own answers on the same set of tasks throughout training process.
We find that this asymmetry still persists: improving a model's generation performance does not lead to corresponding improvements in its ability to verify its own solutions, as illustrated in Figure~\ref{fig:asymmetry_intro} (top).
This naturally raises a key research question: does this asymmetry also manifest in the reverse direction?
In other words, \emph{can improving a model's self-verification ability lead to better generation performance}?

To answer this question, we adopt an alternative training paradigm: instead of training the model to generate better answers, we train it solely to judge the correctness of its own solutions.
With a carefully designed self-verification training pipeline, we surprisingly find that although training the model for generation does not improve its self-verification ability, training the model to self-verify does improve its generation performance, even achieving comparable performance to standard generation training on several benchmarks, as illustrated in Figure~\ref{fig:asymmetry_intro} (down). 
Beyond comparable performance, the resulting models acquire strong verification capability.
Benefiting from this improved self-verification ability, we observe a significant reduction in the number of tokens required to solve the same problems, indicating more efficient reasoning.
Moreover, stronger self-verification unlocks effective test-time scaling: incorporating self-verification results into majority voting leads to performance gains.

Building on these observations, we further explore integrating self-verification into generation training by formulating a multi-task reinforcement learning framework, where generation and self-verification are optimized as two independent but complementary objectives.
Specifically, we introduce two orthogonal training strategies: (i) learning to self-verify as a stronger initial policy before learning to generate, and (ii) alternating training between generation and verification, where a verification phase is triggered after several generation steps.
Extensive experiments show that these integrated training strategies consistently outperform those trained with generation alone. 

Our main contributions are as follows:
\begin{itemize}[leftmargin=*, topsep=2pt, itemsep=0pt]
    \item We conduct an in-depth investigation of the asymmetry between generation and self-verification throughout training, and show that improving generation ability does not lead to corresponding gains in self-verification.
    \item We identify the reverse direction of this asymmetry: learning to self-verify can effectively improve generation performance. Based on this insight, we propose to integrate self-verification into generation training by formulating a multi-task reinforcement learning framework.
    \item We provide extensive experiments demonstrating that learning to self-verify consistently improves problem-solving performance, together with detailed analyses.
\end{itemize}

\section{Preliminary}
\label{sec:preliminary}

\begin{figure*}[t]
    \centering
    \includegraphics[width=0.8\textwidth]{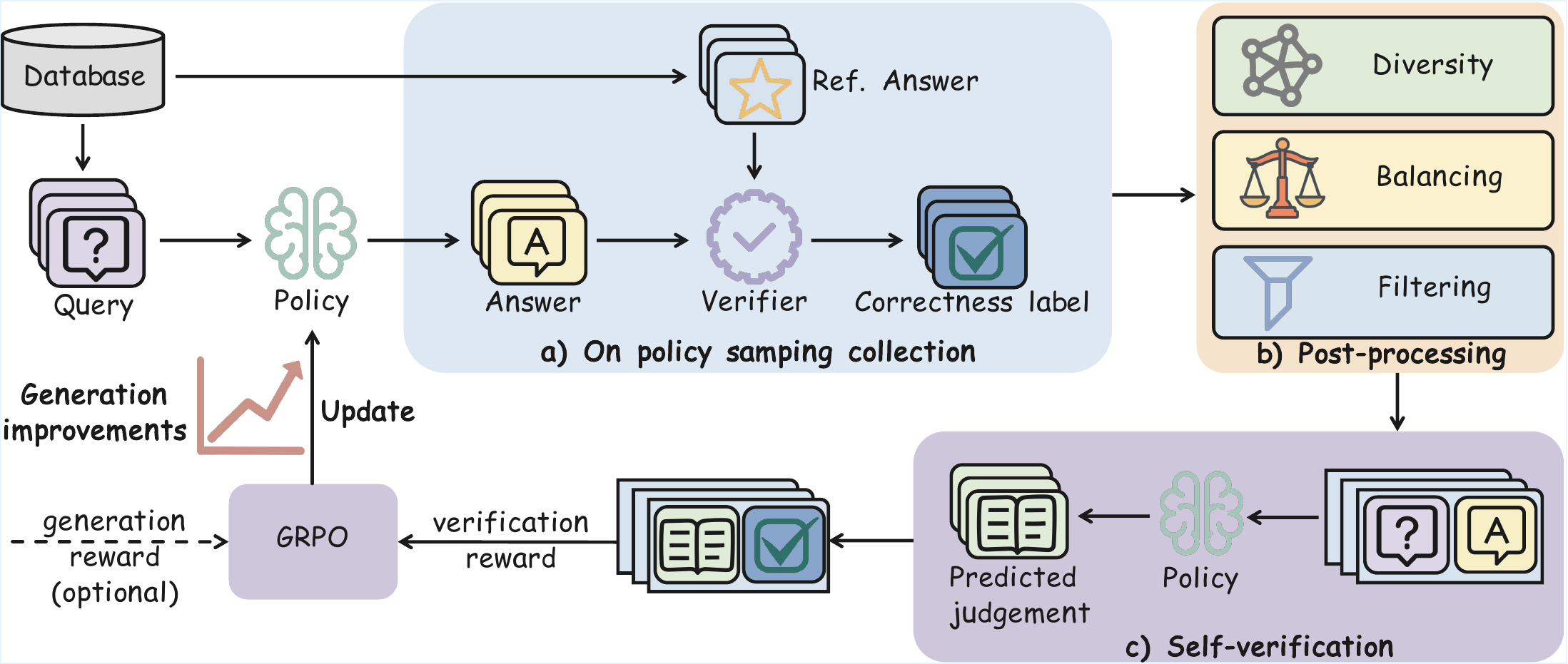}
\caption{
Overview of our self-verification training framework.
We collect on-policy problem-solving trajectories from the model and obtain correctness labels from a verifier.
These trajectories are then processed through a post-processing pipeline, including data balancing, filtering, and diversity-aware sampling, to construct self-verification training data, which is used to train the model to judge the correctness of its own answers.
We find that training the model solely for self-verification already leads to improved generation performance.
Integrating this self-verification objective into generation training further strengthens the model's generation ability.
}
\vspace{-3mm}
\label{fig:method}
\end{figure*}

In this section, we introduce the preliminary concepts and notations used throughout the paper. We first review the RLVR formulation in Section~\ref{subsec:rlvr}. 
We then show how the same RLVR framework can be instantiated in two different settings: generation training (Section~\ref{subsec:generation_training}), where the model is optimized to solve a given task, and verification training (Section~\ref{subsec:verification_training}), where the model is optimized to judge the correctness of a given solution.

\subsection{RLVR}
\label{subsec:rlvr}

Reinforcement Learning with Verifiable Rewards (RLVR) is a reinforcement learning framework for training language models using automatically computable reward signals. 
Instead of relying on human preference models, RLVR employs a rule-based verifier that evaluates each model output against a reference and returns a scalar reward.

Concretely, given an input query $x_i$, where $i$ denotes the query index, the model parameterized by $\pi_\theta$ generates multiple outputs $o_{i,j} = (z_{i,j}, y_{i,j})$, where $j$ indexes different samples generated for the same query. Here, $z_{i,j}$ denotes the intermediate reasoning trace and $y_{i,j}$ denotes the final prediction.
A verifier then assigns a reward score $r_{i,j}$ by comparing the model output with a reference solution $y_i^*$.
The training objective is to optimize the model parameters so as to maximize the expected verifier reward over model-generated samples:
\begin{equation}
\max_{\theta} \; \mathbb{E}_{o \sim \pi_\theta(\cdot \mid x)} \big[ r \big].
\end{equation}

In this work, we adopt Group Relative Policy Optimization (GRPO)~\cite{GRPO} as the underlying optimization algorithm. 
GRPO can be viewed as a simplified variant of PPO~\cite{PPO} that directly optimizes the policy without introducing a separate value network. 
For each input $x_i$, the policy samples a set of $G$ candidate outputs $\{(z_{i,j}, y_{i,j})\}_{j=1}^G$, each receiving a reward $r_{i,j}$.
The policy is then updated by comparing each candidate against the group statistics, using the following clipped surrogate objective:
\begin{equation}
\small
\begin{aligned}
&\mathcal{L}_{\text{GRPO}}(\theta) 
= \mathbb{E}_{x_i \sim \mathcal{D}} \Bigg[ \frac{1}{G} \sum_{j=1}^{G} \frac{1}{|z_{i,j}\circ y_{i,j}|} \\
&\times \sum\limits_{t=1}^{|z_{i,j}\circ y_{i,j}|} \min\Big( \rho^t_{i,j} A_{i,j}, \; \mathrm{clip}(\rho^t_{i,j}, 1 - \epsilon, 1 + \epsilon) A_{i,j} \Big) \Bigg]
\end{aligned}
\end{equation}
where:
\[
\rho_{i,j}^t = \frac{\pi_\theta(z_{i,j}^t\circ y_{i,j}^t \mid x_i, z_{i,j}^{<t}\circ y_{i,j}^{<t})}{\pi_{\theta_{\text{old}}}(z_{i,j}^t\circ y_{i,j}^t \mid x_i, z_{i,j}^{<t}\circ y_{i,j}^{<t})},
\]
where the superscript $t$ denotes the token index in the concatenated sequence, and $\circ$ denotes sequence concatenation.
Here, the advantage $A_{i,j}$ is computed by normalizing the rewards within the sampled group:
\begin{equation}
A_{i,j} = \frac{r_{i,j} - \mu_{i}}{\sigma_{i} + \epsilon_{\text{norm}}},
\end{equation}
with:
\[
\mu_{i} = \frac{1}{G} \sum_{j=1}^{G} r_{i,j}, 
\qquad
\sigma_{i} = \sqrt{ \frac{1}{G} \sum_{j=1}^{G} (r_{i,j} - \mu_{i})^2 }.
\]
Intuitively, GRPO encourages generations that perform better than the group average while suppressing those with lower relative rewards.
Inspired by~\cite{DAPO}, we adopt the clip-higher strategy and token-level mean advantage normalization.

\subsection{Generation Training}
\label{subsec:generation_training}
Under the RLVR formulation, generation training corresponds to the standard task-solving setting. 
Each training example consists of a task query $x_i$ and a reference solution $y_i^*$. 
Given $x_i$, the model samples multiple candidate solutions $\{(z_{i,j}, y_{i,j})\}_{j=1}^G$, and the verifier assigns a reward by checking whether each generated answer $y_{i,j}$ matches the reference solution $y_i^*$.
In this case, the reward signal directly reflects task-solving correctness, and RLVR reduces to optimizing the policy to produce correct solutions for the given tasks.

\subsection{Verification Training}
\label{subsec:verification_training}
Under the same RLVR formulation, verification training corresponds to a different instantiation of the input and reference. 
Each training sample is constructed from a triplet $(x_i, y_{i,j}, c_{i,j})$, where $x_i$ is the task query, $y_{i,j}$ is a candidate solution generated by the model, and $c_{i,j} \in \{0,1\}$ is a binary correctness label indicating whether $y_{i,j}$ matches the reference solution $y_i^*$.
Given such an input $(x_i, y_{i,j})$, the model is prompted to output a judgment $\hat{c}_{i,j}$ indicating whether the provided solution is correct.
A rule-based verifier then assigns a reward by comparing the model's judgment $\hat{c}_{i,j}$ with the reference label $c_{i,j}$.
In this setting, the model is not optimized to solve the task itself, but rather to assess the correctness of given solutions.
\section{Learning to Self-Verify}
\label{sec:method}

\begin{table*}[t]
\caption{
Evaluation of learning to self-verify across six mathematical reasoning benchmarks.
We report both task accuracy (Acc@16 $\uparrow$) and average reasoning length in tokens (Tokens $\downarrow$) for each model trained under two different objectives: train LLM to generate better solutions (Generate), and train LLM to verify its own solutions (Self-Verify).
Results show that models trained with self-verification yield efficient reasoning traces, while achieving comparable or sometimes even better performance than models trained for generation.
}
\centering
\small
\setlength{\tabcolsep}{3pt}

\begin{tabular}{l|cc|cc|cc|cc|cc|cc|cc}
\toprule
\multirow{2}{*}{\textbf{Method}}
& \multicolumn{2}{c}{\textbf{AMC23}}
& \multicolumn{2}{c}{\textbf{Minerva}}
& \multicolumn{2}{c}{\textbf{Olympiad}}
& \multicolumn{2}{c}{\textbf{Math500}}
& \multicolumn{2}{c}{\textbf{AIME24}}
& \multicolumn{2}{c}{\textbf{AIME25}}
& \multicolumn{2}{c}{\textbf{Avg}} \\
\cmidrule(lr){2-3} 
\cmidrule(lr){4-5} 
\cmidrule(lr){6-7} 
\cmidrule(lr){8-9} 
\cmidrule(lr){10-11} 
\cmidrule(lr){12-13}
\cmidrule(lr){14-15}
& Tokens$\downarrow$ & Acc$\uparrow$
& Tokens$\downarrow$ & Acc$\uparrow$
& Tokens$\downarrow$ & Acc$\uparrow$
& Tokens$\downarrow$ & Acc$\uparrow$
& Tokens$\downarrow$ & Acc$\uparrow$
& Tokens$\downarrow$ & Acc$\uparrow$
& Tokens$\downarrow$ & Acc$\uparrow$ \\
\midrule

\multicolumn{15}{c}{\textbf{Qwen2.5-1.5B-Instruct}} \\
\midrule
Generate & 1402 & 30.5 & 963 & 12.4 & 1639 & 20.6 & 936 & 53.7 & 2580 & 2.9 & 2103 & 0.8 & 1604 & 20.2 \\
Self-Verify & 1309 & \textbf{33.0} & 870 & \textbf{14.0} & 1351 & \textbf{22.2} & 817 & \textbf{54.6} & 1467 & \textbf{4.8} & 1545 & \textbf{1.3} & 1227 & \textbf{21.7} \\
\midrule

\multicolumn{15}{c}{\textbf{Qwen2.5-3B-Instruct}} \\
\midrule
Generate & 2754 & \textbf{50.9} & 2006 & 17.0 & 3299 & 27.4 & 2021 & 59.6 & 4811 & 8.1 & 4744 & \textbf{8.1} & 3273 & 28.5 \\
Self-Verify & 1825 & 46.7 & 1658 & \textbf{17.1} & 1891 & \textbf{32.1} & 1237 & \textbf{65.6} & 2755 & \textbf{9.2} & 2252 & 6.3 & 1936 & \textbf{29.5} \\
\midrule

\multicolumn{15}{c}{\textbf{Qwen2.5-7B-Instruct}} \\
\midrule
Generate & 3967 & \textbf{65.3} & 2353 & \textbf{25.3} & 4543 & 37.8 & 2437 & 70.9 & 7053 & 16.3 & 6397 & \textbf{18.1} & 4458 & \textbf{38.9} \\
Self-Verify & 1194 & 59.7 & 823 & 25.2 & 1168 & \textbf{39.8} & 783 & \textbf{74.7} & 1575 & \textbf{19.4} & 1369 & 11.7 & 1152 & 38.4 \\

\bottomrule
\end{tabular}
\label{tab:verify_training}
\end{table*}

In this section, we investigate the reverse direction of this long-standing asymmetry: whether a model can improve its generation performance solely by learning to verify its own solutions.
We first introduce our self-verification training pipeline in Section~\ref{subsec:learn_to_self_verify}, then describe the experimental setup in Section~\ref{subsec:setup_self_verify}, present the main results in Section~\ref{subsec:result_self_verify}, and provide further analysis in Section~\ref{subsec:analysis_self_verify}.

\subsection{Self-Verification Framework}
\label{subsec:learn_to_self_verify}
Following the notation in Section~\ref{sec:preliminary}, we now introduce our self-verification framework, as illustared in Figure~\ref{fig:method}.
\paragraph{On-Policy Sample Collection}
At each training iteration, we sample a mini-batch of $B$ queries $\{x_i\}_{i=1}^B$. 
For each query $x_i$, we use the current policy $\pi_\theta$ to generate $G$ candidate answers, resulting in $B \times G$ generated samples.
For each generated sample, the model produces a solution $y_{i,j}$ together with its corresponding reasoning trace $z_{i,j}$, where $j = 1, \ldots, G$. 
A rule-based verifier then compares each $y_{i,j}$ with the reference answer $y_i^*$ and assigns a binary correctness label $c_{i,j}$.
Each sample is thus represented as a triplet $(x_i, y_{i,j}, c_{i,j})$.
All such triplets are stored in a temporary buffer and serve as the raw candidates for constructing the self-verification training data.

\paragraph{Post-Processing}
At each iteration, the on-policy sampling procedure produces $B \times G$ samples. Directly using all of them for verification training is computationally expensive and can also introduce instability due to imbalance or low-quality samples. For a fair comparison, we downsample these candidates and construct a verification training batch of size $B$ by selecting the most informative samples.
Specifically, we apply the following steps:
\begin{itemize}[leftmargin=*, topsep=2pt, itemsep=0pt]
    \item \textbf{Filtering:} We first discard invalid samples, including those with malformed outputs, excessively long generations, or missing a unique final answer. We further discard queries for which all generated answers are incorrect, as such cases typically exceed the current capability of the model and provide little useful supervision signal for self-verification.
    \item \textbf{Diversity Control:} To avoid overfitting to a small subset of queries when conducting self-verification training, we perform sampling at the query level and ensure that the selected verification samples are drawn from diverse input queries.
    \item \textbf{Data Balancing:} Since generation often produces highly imbalanced labels (e.g., mostly incorrect at early stages and mostly correct at later stages), while self-verification is essentially a binary classification task, we explicitly enforce each mini-batch of verification data to contain an equal number of correct and incorrect samples.
\end{itemize}

\paragraph{Training}
In self-verification training, the model is prompted with a query--answer pair $(x_i, y_{i,j})$ and is required to predict whether the provided answer is correct.
Let $\hat{c}_{i,j}$ denote the model's predicted judgment and $c_{i,j}$ the reference correctness label obtained from the rule-based verifier.
A verification reward is then computed as:
\begin{equation}
r_{i,j}^v = \mathrm{Verifier}(\hat{c}_{i,j}, c_{i,j}).
\end{equation}
We then optimize the model using the same GRPO objective as in Section~\ref{subsec:rlvr}.
This training stage treats the model purely as a verifier and encourages it to improve its ability to distinguish correct from incorrect answers. We emphasize that at this stage, the training objective contains no generation reward. The policy is optimized solely to maximize the expected verification reward.

\begin{figure}[t]
    \centering
    \includegraphics[width=0.48\textwidth]{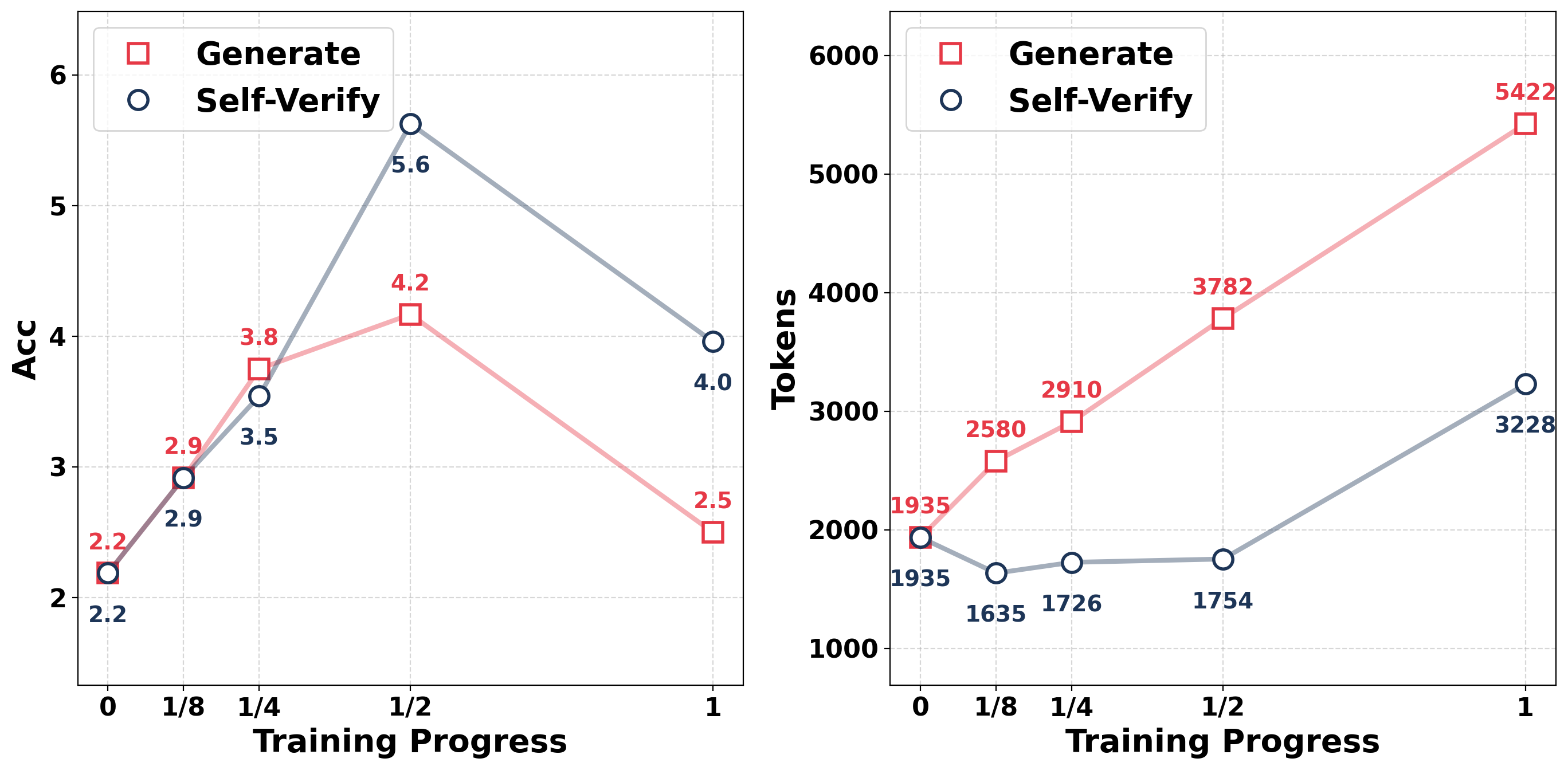}
    \vspace{-1.5em}
    \caption{
Comparison of accuracy and token usage between generation training and self-verification training on AIME24 with Qwen2.5-1.5B-Instruct.
    }
\label{fig:training-dynamic}
\end{figure}
\subsection{Experimental Setup}
\label{subsec:setup_self_verify}

We conduct extensive experiments to compare the effects of training LLMs to generate solutions and training them to self-verify. We first describe our experimental setup.

\paragraph{Dataset and Benchmarks}
For training, we use DAPO-Math-17K~\cite{DAPO}, a dataset widely adopted for mathematical reasoning. 
We evaluate our models on six challenging mathematical reasoning benchmarks: AIME24~\cite{aime24}, AIME25~\cite{aime25}, AMC23, Minerva, MATH500~\cite{Math500}, and OlympiadBench~\cite{OlympiaBench}.

\paragraph{Implementation}
We choose Qwen2.5-1.5B-Instruct, Qwen2.5-3B-Instruct, and Qwen2.5-7B-Instruct as backbone models~\cite{Qwen2.5}. 
We use veRL~\cite{verl} as the training framework to implement our RL-based methods with a rule-based verifier. 
For both generation training and self-verification training, we train the models for 1000 steps.
For fair comparison, both generation and verification training use a batch size of 128 with a group size of 8.
We set the maximum generation length to 10,240 tokens for all models, with the temperature set to 0.6 and top-$p$ to 0.95.

\paragraph{Evaluation} 
We compare models trained exclusively for generation with those trained exclusively for self-verification on the benchmarks. 
We report two main metrics: (1) \textbf{Acc}, measured by Avg@16 accuracy, (2) \textbf{Token}, calculated as the average number of tokens (including both intermediate reasoning and the final answer) across all outputs on each test set. This metric reflects the reasoning efficiency of the model.

\subsection{Main Results}
\label{subsec:result_self_verify}
Table~\ref{tab:verify_training} summarizes the performance and reasoning length across six benchmarks and three models, comparing models trained solely to generate answers with models trained solely to judge the correctness of their own solutions.
In addition, Figure~\ref{fig:training-dynamic} illustrates the evolution of accuracy and token usage throughout training on AIME24 with Qwen2.5-1.5B-Instruct.
From the results, we can draw two key conclusions:

\paragraph{Learning to self-verify achieves comparable performance to learning to generate.}
Across all models and datasets, training the model solely for self-verification yields performance that is comparable to, and in some cases better than, that achieved by generation-only training.
For example, for Qwen2.5-1.5B-Instruct, the self-verification-trained model outperforms the generation-trained model in accuracy across all benchmarks.
For Qwen2.5-3B-Instruct, self-verification achieves 32.1\% accuracy on OlympiadBench and 65.6\% on Math500, surpassing the generation baseline by 4.7\% and 6.0\%, respectively, demonstrating strong potential even without explicit generation training.
This points to an interesting asymmetry in the reverse direction: while improving a model's generation performance does not lead to a corresponding improvement in its ability to self-verify, even on the same task (\emph{cf.}\ Figure~\ref{fig:asymmetry_intro}), improving self-verification alone can in turn enhance generation performance.

\begin{table}[t]
  \centering
  \caption{
Evaluation of verification capability.
We report the Acc@8 of different models in judging the correctness of solutions generated by DeepSeek-R1-Distill-Qwen-7B.
  }
  \small
  \resizebox{\columnwidth}{!}{
    \begin{tabular}{cccc}
      \toprule
      \textbf{Model} & \textbf{Base} & \textbf{Generate} & \textbf{Self-Verify} \\
      \midrule
      Qwen2.5-1.5B-Instruct & 45.58 & 45.95$^{\textcolor{mygreen}{+0.37}}$ & \textbf{62.31}$^{\textcolor{mygreen}{+16.73}}$ \\
      Qwen2.5-3B-Instruct   & 59.82 & 55.19$^{\textcolor{myred}{-4.63}}$ & \textbf{65.69}$^{\textcolor{mygreen}{+5.87}}$ \\
      Qwen2.5-7B-Instruct   & 64.46 & 68.84$^{\textcolor{mygreen}{+4.38}}$ & \textbf{69.50}$^{\textcolor{mygreen}{+5.04}}$ \\
      \bottomrule
    \end{tabular}
  }
\label{table:verification} 
\end{table}

\paragraph{Learning to self-verify requires significantly fewer tokens to solve the same problems.}
Across all models and datasets, training the model solely for self-verification consistently produces much shorter reasoning traces than generation-only training, while maintaining comparable performance.
Notably, for Qwen2.5-7B-Instruct, the self-verification-trained model achieves performance comparable to generation training using only about 25\% of the tokens.
For Qwen2.5-3B-Instruct, it uses roughly 60\% of the tokens while even slightly outperforming the generation baseline.
These results indicate that, although the final performance is comparable, self-verification leads to substantially more efficient reasoning traces.
We attribute this to the strengthened self-verification ability induced by our training, which enables the model to better recognize when its current solution is likely incorrect and when verification should be triggered.
As a result, the model avoids redundant or ``fake'' verification behaviors and follows more direct solution trajectories.
This markedly different reasoning behavior further motivates us to regard self-verification and generation as complementary training signals.

\begin{figure}[h]
    \centering
    \includegraphics[width=0.4\textwidth]{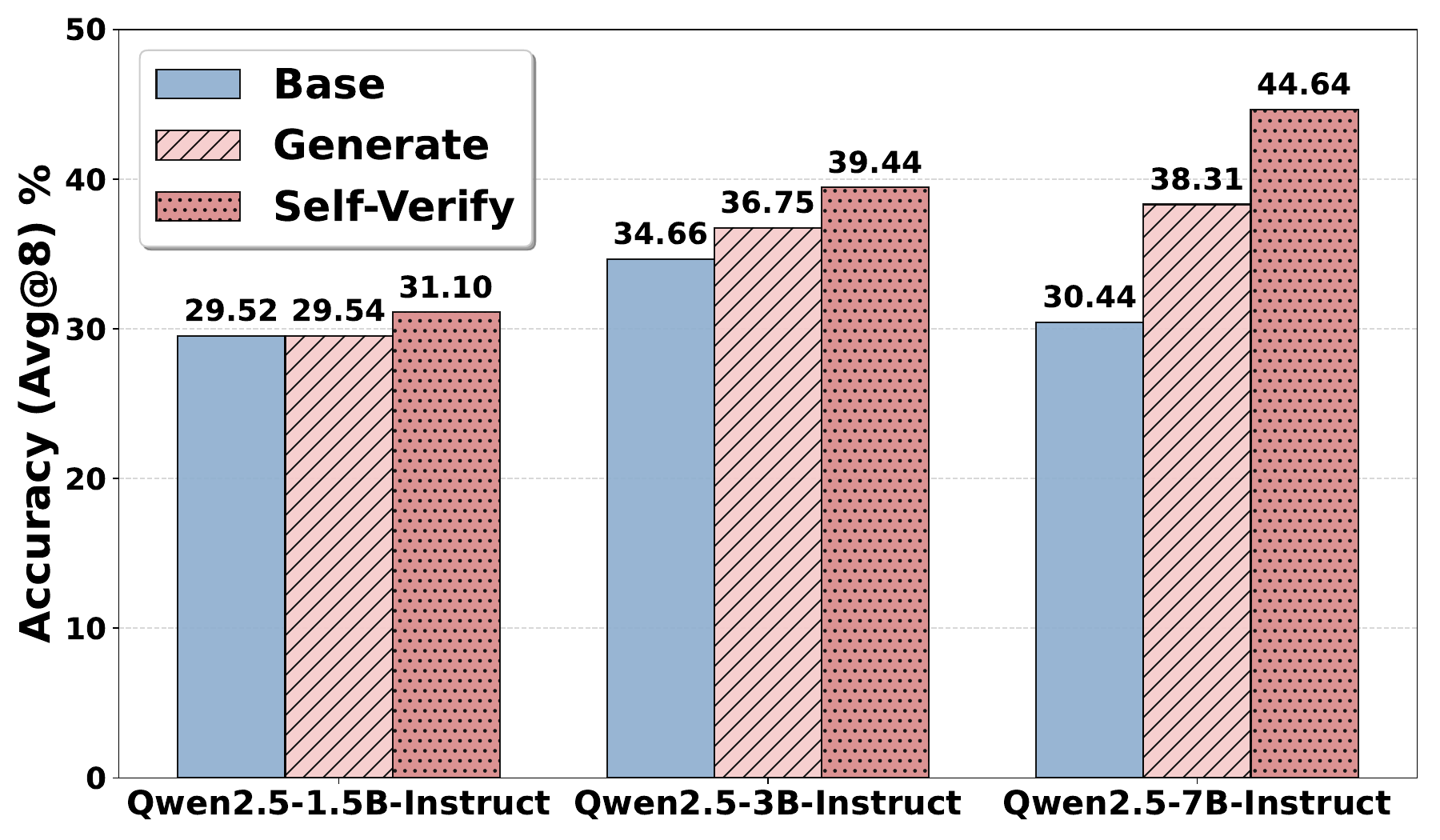}
    \vspace{-1.5em}
    \caption{
Performance comparison under partially corrupted reasoning prefix setting.
}
\label{fig:error-prefix}
\end{figure}

\subsection{Analysis}
\label{subsec:analysis_self_verify}
In this section, we conduct a detailed analysis to investigate what capabilities are acquired by learning to self-verify and how these capabilities can be exploited in practice. 
Based on our experiments, we make the following observations:

\paragraph{Explicit self-verification training turns the model into a strong verifier.}
Figure~\ref{fig:asymmetry_intro} demonstrates that even models with limited parameter sizes can verify their own solutions much more accurately after self-verification training. 
To further evaluate the model's verification capability as a general verifier in specific domains, we construct a verification evaluation set consisting of benchmark solutions generated by DeepSeek-R1-Distill-Qwen-7B. 
The model is then asked to judge whether each solution is correct or incorrect, and the results are reported in Table~\ref{table:verification}.
The results show that, since self-verification is essentially a verification task, our model naturally acquires the ability to assess solutions produced by other models as well, demonstrating strong general-purpose verification capability.
In contrast, models trained only for generation overall achieve marginal performance gain and in some cases even exhibit noticeable degradation.

\paragraph{Learning to self-verify enables the model to identify and correct errors in its reasoning process.}
We observe that after self-verification training, the number of tokens required to solve a problem is significantly reduced. 
This suggests that the model starts to precisely trigger verification when it detects potential errors in its reasoning and to correct them in time, thereby avoiding redundant or ``fake'' verification behaviors.
To validate this hypothesis, we construct a dedicated evaluation set to assess the effectiveness of self-verification behaviors during the reasoning process. 
Specifically, we mix data from different benchmarks and build a set of 1,545 problems. 
We first collect the original reasoning trajectories generated by Qwen2.5-7B-Instruct on these problems. 
We then use GPT-4.1~\cite{GPT4} to randomly rewrite these trajectories into reasoning step prefixes with varying numbers of steps and injected some mistakes. 
The model is prompted with the original query and the corrupted prefix, and is asked to continue the reasoning process.
Under this setting, a higher success rate indicates that the model is more capable of detecting errors in the ongoing reasoning and correcting them through effective self-verification. 
As shown in Figure~\ref{fig:error-prefix}, the self-verification-trained model significantly outperforms both the base model and the generation-trained model, demonstrating substantially stronger error detection and correction capability during reasoning. 
In contrast, generation training yields only marginal improvements over the base model in this setting.

\begin{table}[t]
  \centering
  \caption{
Test-time scaling with self-verification with Qwen2.5-1.5B-Instruct.
We report Acc@32 and compare standard majority voting and majority voting augmented with self-verification across three training regimes: Base, Generate, and Self-Verify.
  }
  
    \resizebox{\columnwidth}{!}{\begin{tabular}{ccccc}
      \toprule
      \textbf{Method} & \textbf{AIME25} & \textbf{MATH500} & \textbf{Olympaid} &
      \textbf{Minerva} \\
      \midrule
      
      \multicolumn{5}{c}{\textbf{Base}} \\ 
      \midrule
      Major voting   & 3.30  & 52.20 & 22.40 & 14.30 \\
      + Self-verify  & 3.30$^{\color{color+}+0.00}$  & 52.40$^{\color{color+}+0.20}$ & 22.30$^{\color{color-}-0.10}$ & 10.70$^{\color{color-}-3.60}$\\

      \midrule
      \multicolumn{5}{c}{\textbf{Generate}} \\ 
      \midrule
      Major voting   & 0.00  & 54.20 & 23.40 & 15.80 \\
      + Self-verify  & 0.00$^{\color{color+}+0.00}$  & 53.00$^{\color{color-}-1.20}$ & 23.60$^{\color{color+}+0.20}$ & 13.60$^{\color{color-}-2.20}$ \\
      
      \midrule
      \multicolumn{5}{c}{\textbf{Self-Verify}} \\ 
      \midrule
      Major voting   & 3.30  & 55.20 & 25.80 & 16.20 \\
      + Self-verify  & \textbf{6.70}$^{\color{color+}+3.40}$ & \textbf{56.40}$^{\color{color+}+1.20}$ & \textbf{27.20}$^{\color{color+}+1.40}$ &
      \textbf{16.20}$^{\color{color+}+0.00}$ \\
      
      \bottomrule
    \end{tabular}}
\label{table:tts}
\end{table}

\paragraph{Effective self-verification enables test-time scaling.}
With a substantially improved self-verification capability, the model can reliably assess the correctness of its own candidate solutions, which unlocks a new form of test-time scaling based on self-verification.
Specifically, at inference time, we sample multiple candidate solutions, let the model verify each of them, and aggregate the verification results to obtain a verification score for each candidate.
We then jointly consider the majority vote and the verification scores to determine the final answer.
Experimental results in Table~\ref{table:tts} show that introducing this additional self-verification signal at test time consistently improves performance, demonstrating that self-verification provides an effective and principled way to scale inference beyond naive sampling or self-consistency.

\begin{table*}[t]
\caption{Evaluation of integrating self-verification into generation training across six mathematical reasoning benchmarks. We report task accuracy (Acc@16 $\uparrow$) for each model under four training strategies.
Results show that our strategies improve overall performance over standard and mixed training.}
\centering
\small

\setlength{\tabcolsep}{10pt}

\begin{tabular}{cccccccc}
\toprule
\textbf{Method}
& \textbf{AMC23}
& \textbf{Minerva}
& \textbf{Olympiad}
& \textbf{Math500}
& \textbf{AIME24}
& \textbf{AIME25}
& \textbf{Avg} \\
\midrule

\multicolumn{8}{c}{\textbf{Qwen2.5-1.5B-Instruct}} \\
\midrule
Generate & 30.5 & 12.4 & 20.6 & 53.7 & 2.9 & 0.8 & 20.2 \\
Mixed-Train & 33.3 & 12.7 & 21.2 & 53.8 & 3.8 & 1.5 & 21.1 \\
Verify-Init & 33.0 & 13.0 & 21.9 & \textbf{54.7} & \textbf{5.4} & 1.3 & 21.6 \\
Verify-Alter & \textbf{36.4} & \textbf{13.9} & \textbf{22.4} & 54.2 & 5.0 & \textbf{4.2} & \textbf{22.7} \\
\midrule

\multicolumn{8}{c}{\textbf{Qwen2.5-3B-Instruct}} \\
\midrule
Generate & \textbf{50.9} & 17.0 & 27.4 & 59.6 & 8.1 & \textbf{8.1} & 28.5 \\
Mixed-Train & 49.4 & 17.6 & 27.5 & 59.2 & \textbf{10.8} & 6.3 & 28.5 \\
Verify-Init & 47.8 & 18.3 & 29.5 & 63.1 & 9.6 & 6.5 & 29.1 \\
Verify-Alter & 47.7 & \textbf{18.7} & \textbf{30.2} & \textbf{64.5} & 9.4 & 5.6 & \textbf{29.4} \\
\midrule

\multicolumn{8}{c}{\textbf{Qwen2.5-7B-Instruct}} \\
\midrule
Generate & 65.3 & 25.3 & 37.8 & 70.9 & 16.3 & \textbf{18.1} & 38.9 \\
Mixed-Train & 59.2 & 24.6 & \textbf{40.5} & 73.5 & 15.8 & 9.8 & 37.2 \\
Verify-Init & 63.6 & \textbf{26.0} & 39.0 & \textbf{74.0} & 17.3 & 12.5 & 38.7 \\
Verify-Alter & \textbf{68.0} & \textbf{26.0} & 39.0 & 72.9 & \textbf{18.3} & 17.7 & \textbf{40.3} \\

\bottomrule
\end{tabular}
\label{tab:integrate_training}
\end{table*}

\section{Integrating Self-Verification into Training}
We observe that training a model solely to verify its own answers already improves its generation performance to a level comparable with models trained purely for generation, while exhibiting a markedly different inference behavior: verification-only models produce significantly shorter outputs, indicating a more efficient reasoning trace. This markedly different reasoning behavior further motivates us to view self-verification and generation as complementary training signals.
Building on this observation, we further propose to integrate self-verification into generation training.

\subsection{Multi-Task RL Pipeline}
In this work, we formulate the integration of generation and self-verification as a multi-task reinforcement learning problem, where the two objectives are optimized in a decoupled manner. 
Under this framework, we consider two simple yet effective strategies that are orthogonal: a stage-wise initialization strategy and an alternating training strategy.

\paragraph{Stage-wise Initialization}
We first train the model with a self-verification objective by optimizing the policy to maximize the verification reward $r_v$, as described in Section~\ref{subsec:learn_to_self_verify}. 
The resulting model, which already possesses a stronger ability to judge the correctness of its own outputs, is then used as a better initial policy for standard generation training, where the policy is further optimized to maximize the generation reward $r_g$. 

\paragraph{Alternating Training}
We alternate between generation training and self-verification training. 
Specifically, we run generation training for $n$ steps to optimize the policy with respect to the generation reward $r_g$. Every $n$ generation steps, we trigger a self-verification phase, during which the same policy is optimized with respect to the verification reward $r_v$, using the answers generated in the preceding generation phase to construct verification training data.
This process is repeated throughout training, allowing the policy to be continuously shaped by both objectives.

In both strategies, generation and self-verification are optimized under the same RLVR framework using GRPO, and the only difference lies in which reward signal ($r_g$ or $r_v$) is used at each stage of training.

\subsection{Experimental Setup}

\paragraph{Baseline}
To benchmark the effectiveness of our method, we compare it against two primary baselines. 
\emph{Generate} follows the standard RL-based training paradigm for reasoning models and optimizes the policy solely with respect to the generation reward. 
\emph{Mixed-Train}~\cite{join_train2} jointly optimizes generation and self-verification objectives within each training step.
For fair comparison, all baselines and our methods are trained using the same implementation and the same set of hyperparameters.

\paragraph{Implementation and Evaluation}
We use the same datasets, model architectures, implementation details, and evaluation protocols as in Section~\ref{subsec:result_self_verify}.
In this section, we evaluate two strategies for integrating self-verification into training: \emph{Verify-Init}, which corresponds to the stage-wise initialization strategy, and \emph{Verify-Alter}, which corresponds to the alternating training strategy.
For \emph{Verify-Init}, we initialize the model from a checkpoint obtained after 400 steps of self-verification-only training, and then further train it for 600 steps with the generation objective.
For \emph{Generate}, \emph{Mixed-Train}, and \emph{Verify-Alter}, we train the models for 1000 steps in total.

\subsection{Results and Analysis}
Table~\ref{tab:integrate_training} summarizes the performance across six benchmarks and three models.
Beyond training the model solely for self-verification, we find that integrating self-verification into generation training consistently leads to improved generation performance across most models and benchmarks.
Compared to \emph{Mixed-Train}, which directly mixes the two objectives within a single optimization step, our framework decouples the optimization of generation and verification and optimizes them in a coordinated but separate manner, demonstrating additional performance gains.
For instance, for Qwen2.5-1.5B-Instruct, \emph{Verify-Alter} improves the average accuracy from 20.2\% to 22.7\%, outperforming both standard generation training and mixed-objective training.
Notably, on AMC23, it improves accuracy by 5.9 points, and on the more challenging AIME benchmarks, it raises accuracy from 0.8\% to 4.2\%.
This suggests that self-verification provides a complementary and beneficial training signal.

\section{Related Works}

\subsection{LLM as Generator}
Improving the generation capability of LLMs has long been a central focus of the community~\cite{GPT3, GPT4}. 
Early works typically collect high-quality trajectories with complex reasoning patterns and train LLMs via imitation learning~\cite{InstructGPT, llama1, Qwen2.5}. 
With the success of models such as DeepSeek-R1~\cite{Deepseek-R1}, which is trained with the GRPO algorithm~\cite{GRPO}, a surge of follow-up research has been inspired. 
Meanwhile, Reinforcement Learning with Verifiable Rewards (RLVR) has emerged as a powerful and scalable training paradigm for further boosting LLM generation performance by leveraging verifiable reward signals~\cite{search-r1, ragen}.
Building on these foundations, more recent studies start to investigate how to extend LLMs' advanced generation ability to broader domains~\cite{diverse1, diverse2, diverse3}, make generation more efficient at inference time~\cite{efficient1, efficient2, efficient3}, and improve stability and effectiveness of generation training~\cite{stable1, effective1, effective2}. 
These advances have led to a series of increasingly capable large models~\cite{GPT-5.2, Claude-Opus-4.5, Gemini3}. 
However, despite these advancements, even the most powerful LLMs still cannot reliably self-verify their own outputs~\cite{cant_scale, cant_self_verify1}.

\subsection{LLMs as Verifier}
Verifiers play a crucial role in guiding LLMs toward better generations and enabling effective test-time scaling~\cite{reward_model_survey, reward_model_survey2, test_time_scaling}. 
Existing verifiers are typically either (1) discriminative~\cite{scalar_reward1, scalar_reward2}, producing scalar scores to rank candidate responses, or (2) generative~\cite{GRM1, GRM2, GRM3}, producing textual judgments or reward signals. 
With the success of RLVR in training stronger generators, increasing attention has been paid to generative verifiers due to their better generalization ability. 
LLM verifiers typically produce natural language rationales or textual judgments, which improve transparency and evaluation reliability. 
Correspondingly, training methods for LLM verifiers have evolved from supervised fine-tuning (SFT) to direct preference optimization (DPO)~\cite{rm-dpo, GRM1, GRM3}, and more recently to RLVR~\cite{rm-r1, rewardanything}, inspired by advances in reasoning-oriented models.
Despite these advances, how training as verifiers influences the model itself as a generator remains largely underexplored.

\subsection{Joint Training of Generator and Verifier}
Recently, several works have begun to explore incorporating verification signals into generator training. 
Among them, \cite{verithinker} shows that collecting correctness signals on external models' outputs and training LLMs via imitation learning with fixed templates can shorten generated responses, albeit sometimes at the cost of slightly degraded generation performance. 
Other works~\cite{join_train1, join_train2, join_train3} propose to jointly train generation and verification within the same training step, where verification is scaled as an auxiliary signal while the overall training dynamics remain dominated by the generation objective.

Different from these approaches, we notice that rewarding self-verification alone is sufficient to obtain a generator with performance comparable to standard generation training, while producing better reasoning traces. 
We further explore integrating self-verification into generation training by formulating a multi-task reinforcement learning framework, where generation and self-verification are optimized as two decoupled but complementary objectives.

\section{Conclusion}
In this work, we investigate the asymmetry between generation and self-verification in large language models and show that improving generation does not naturally lead to better self-verification, even on the same task. 
More interestingly, we identify the reverse direction of this asymmetry: learning to self-verify alone can significantly improve generation performance. 
This finding challenges the common view of verification as merely an auxiliary component and highlights its role as a powerful training signal.
Building on this insight, we further explore integrating self-verification into generation training by formulating a multi-objective reinforcement framework.
Extensive experiments demonstrate that explicit self-verification training consistently improves problem-solving performance, produces more efficient and effective reasoning traces, and enables effective test-time scaling.
Looking ahead, we believe that verification has untapped potential to improve generation, through designed verification tasks, more principled integration of verification and generation objectives, and more efficient training strategies.
Exploring these directions is beyond the scope of the current work, and we leave them for future research.

\section*{Impact Statement}
Our findings suggest that strengthening self-verification in large language models can fundamentally change how these systems reason and generate responses. By showing that learning to self-verify not only improves reliability but also enhances generation efficiency, this work contributes to a better understanding of the interaction between reasoning, verification, and generation in modern language models.
These insights have broader implications for the development and deployment of AI systems, especially in scenarios where correctness, robustness, and controllability are critical. Improving a model’s ability to assess its own outputs may help reduce spurious reasoning steps, increase transparency, and mitigate certain classes of errors in real-world applications. At the same time, more powerful self-verification capabilities also raise new questions about how such systems should be evaluated, monitored, and governed, particularly when they are used in high-stakes or decision-critical settings. Understanding and carefully managing these dynamics is therefore important for the responsible use of large language models in practice.

\section*{Limitation}
\label{limitation}
Although this work provides an encouraging analysis of the asymmetry between generation and self-verification and demonstrates the effectiveness of learning to self-verify, it still has several limitations.
First, introducing an additional self-verification objective into generation training inevitably incurs extra computation, including additional inference and optimization costs. 
Second, although our experiments cover models of different parameter sizes, they are still limited in scale. 
Due to computational constraints, we do not explore whether the same phenomena and benefits continue to hold for larger models.
Also, despite its effectiveness, we only explore a limited set of ways to combine generation and self-verification, as well as a single form of verification task. More diverse verification formulations and tighter coupling paradigms between generation and verification may further push the performance ceiling.
Moreover, our study focuses primarily on mathematical reasoning benchmarks. While the proposed framework is conceptually general, it remains an open question whether the same asymmetry and the benefits of learning to self-verify will hold in other domains, such as planning and multimodal reasoning.
Finally, the current multi-task training schedule (e.g., stage-wise or alternating) is manually designed and heuristic. A more principled or adaptive strategy for balancing generation and self-verification objectives remains an interesting direction for future work.

\bibliography{reference}

@article{Deepseek-R1,
  author       = {DeepSeek{-}AI},
  title        = {DeepSeek-R1: Incentivizing Reasoning Capability in LLMs via Reinforcement
                  Learning},
  journal      = {CoRR},
  volume       = {abs/2501.12948},
  year         = {2025}
}

@article{Qwen3,
  author       = {An Yang and
                  Anfeng Li and
                  Baosong Yang and
                  Beichen Zhang and
                  Binyuan Hui and
                  Bo Zheng and
                  Bowen Yu and
                  Chang Gao and
                  Chengen Huang and
                  Chenxu Lv and
                  Chujie Zheng and
                  Dayiheng Liu and
                  Fan Zhou and
                  Fei Huang and
                  Feng Hu and
                  Hao Ge and
                  Haoran Wei and
                  Huan Lin and
                  Jialong Tang and
                  Jian Yang and
                  Jianhong Tu and
                  Jianwei Zhang and
                  Jian Yang and
                  Jiaxi Yang and
                  Jingren Zhou and
                  Junyang Lin and
                  Kai Dang and
                  Keqin Bao and
                  Kexin Yang and
                  Le Yu and
                  Lianghao Deng and
                  Mei Li and
                  Mingfeng Xue and
                  Mingze Li and
                  Pei Zhang and
                  Peng Wang and
                  Qin Zhu and
                  Rui Men and
                  Ruize Gao and
                  Shixuan Liu and
                  Shuang Luo and
                  Tianhao Li and
                  Tianyi Tang and
                  Wenbiao Yin and
                  Xingzhang Ren and
                  Xinyu Wang and
                  Xinyu Zhang and
                  Xuancheng Ren and
                  Yang Fan and
                  Yang Su and
                  Yichang Zhang and
                  Yinger Zhang and
                  Yu Wan and
                  Yuqiong Liu and
                  Zekun Wang and
                  Zeyu Cui and
                  Zhenru Zhang and
                  Zhipeng Zhou and
                  Zihan Qiu},
  title        = {Qwen3 Technical Report},
  journal      = {CoRR},
  volume       = {abs/2505.09388},
  year         = {2025}
}

@article{Gemma3,
  author       = {Gemma Team},
  title        = {Gemma 3 Technical Report},
  journal      = {CoRR},
  volume       = {abs/2503.19786},
  year         = {2025}
}

@misc{Gemini3,
title = {A new era of intelligence with Gemini 3},
author = {Google},
year = {2025},
url = {https://blog.google/products-and-platforms/products/gemini/gemini-3/#note-from-ceo}
}

@misc{GPT-5.2,
title = {Introducing GPT-5.2},
author = {OpenAI},
year = {2025},
url = {https://openai.com/index/introducing-gpt-5-2/}
}

@article{deepseek-math-v2,
  author       = {Zhihong Shao and
                  Yuxiang Luo and
                  Chengda Lu and
                  Z. Z. Ren and
                  Jiewen Hu and
                  Tian Ye and
                  Zhibin Gou and
                  Shirong Ma and
                  Xiaokang Zhang},
  title        = {DeepSeekMath-V2: Towards Self-Verifiable Mathematical Reasoning},
  journal      = {CoRR},
  volume       = {abs/2511.22570},
  year         = {2025}
}

@misc{Claude-Opus-4.5,
title = {Introducing Claude Opus 4.5},
author = {Anthropic},
year = {2025},
url = {https://www.anthropic.com/news/claude-opus-4-5}
}

@misc{GLM-4.7,
title = {GLM-4.7: Advancing the Coding Capability},
author = {Z.AI},
year = {2025},
url = {https://z.ai/blog/glm-4.7}
}

@misc{DeepSeek-V3.2,
      title={DeepSeek-V3.2: Pushing the Frontier of Open Large Language Models}, 
      author={DeepSeek-AI and Aixin Liu and Aoxue Mei and Bangcai Lin and Bing Xue and Bingxuan Wang and Bingzheng Xu and Bochao Wu and Bowei Zhang and Chaofan Lin and Chen Dong and Chengda Lu and Chenggang Zhao and Chengqi Deng and Chenhao Xu and Chong Ruan and Damai Dai and Daya Guo and Dejian Yang and Deli Chen and Erhang Li and Fangqi Zhou and Fangyun Lin and Fucong Dai and Guangbo Hao and Guanting Chen and Guowei Li and H. Zhang and Hanwei Xu and Hao Li and Haofen Liang and Haoran Wei and Haowei Zhang and Haowen Luo and Haozhe Ji and Honghui Ding and Hongxuan Tang and Huanqi Cao and Huazuo Gao and Hui Qu and Hui Zeng and Jialiang Huang and Jiashi Li and Jiaxin Xu and Jiewen Hu and Jingchang Chen and Jingting Xiang and Jingyang Yuan and Jingyuan Cheng and Jinhua Zhu and Jun Ran and Junguang Jiang and Junjie Qiu and Junlong Li and Junxiao Song and Kai Dong and Kaige Gao and Kang Guan and Kexin Huang and Kexing Zhou and Kezhao Huang and Kuai Yu and Lean Wang and Lecong Zhang and Lei Wang and Liang Zhao and Liangsheng Yin and Lihua Guo and Lingxiao Luo and Linwang Ma and Litong Wang and Liyue Zhang and M. S. Di and M. Y Xu and Mingchuan Zhang and Minghua Zhang and Minghui Tang and Mingxu Zhou and Panpan Huang and Peixin Cong and Peiyi Wang and Qiancheng Wang and Qihao Zhu and Qingyang Li and Qinyu Chen and Qiushi Du and Ruiling Xu and Ruiqi Ge and Ruisong Zhang and Ruizhe Pan and Runji Wang and Runqiu Yin and Runxin Xu and Ruomeng Shen and Ruoyu Zhang and S. H. Liu and Shanghao Lu and Shangyan Zhou and Shanhuang Chen and Shaofei Cai and Shaoyuan Chen and Shengding Hu and Shengyu Liu and Shiqiang Hu and Shirong Ma and Shiyu Wang and Shuiping Yu and Shunfeng Zhou and Shuting Pan and Songyang Zhou and Tao Ni and Tao Yun and Tian Pei and Tian Ye and Tianyuan Yue and Wangding Zeng and Wen Liu and Wenfeng Liang and Wenjie Pang and Wenjing Luo and Wenjun Gao and Wentao Zhang and Xi Gao and Xiangwen Wang and Xiao Bi and Xiaodong Liu and Xiaohan Wang and Xiaokang Chen and Xiaokang Zhang and Xiaotao Nie and Xin Cheng and Xin Liu and Xin Xie and Xingchao Liu and Xingkai Yu and Xingyou Li and Xinyu Yang and Xinyuan Li and Xu Chen and Xuecheng Su and Xuehai Pan and Xuheng Lin and Xuwei Fu and Y. Q. Wang and Yang Zhang and Yanhong Xu and Yanru Ma and Yao Li and Yao Li and Yao Zhao and Yaofeng Sun and Yaohui Wang and Yi Qian and Yi Yu and Yichao Zhang and Yifan Ding and Yifan Shi and Yiliang Xiong and Ying He and Ying Zhou and Yinmin Zhong and Yishi Piao and Yisong Wang and Yixiao Chen and Yixuan Tan and Yixuan Wei and Yiyang Ma and Yiyuan Liu and Yonglun Yang and Yongqiang Guo and Yongtong Wu and Yu Wu and Yuan Cheng and Yuan Ou and Yuanfan Xu and Yuduan Wang and Yue Gong and Yuhan Wu and Yuheng Zou and Yukun Li and Yunfan Xiong and Yuxiang Luo and Yuxiang You and Yuxuan Liu and Yuyang Zhou and Z. F. Wu and Z. Z. Ren and Zehua Zhao and Zehui Ren and Zhangli Sha and Zhe Fu and Zhean Xu and Zhenda Xie and Zhengyan Zhang and Zhewen Hao and Zhibin Gou and Zhicheng Ma and Zhigang Yan and Zhihong Shao and Zhixian Huang and Zhiyu Wu and Zhuoshu Li and Zhuping Zhang and Zian Xu and Zihao Wang and Zihui Gu and Zijia Zhu and Zilin Li and Zipeng Zhang and Ziwei Xie and Ziyi Gao and Zizheng Pan and Zongqing Yao and Bei Feng and Hui Li and J. L. Cai and Jiaqi Ni and Lei Xu and Meng Li and Ning Tian and R. J. Chen and R. L. Jin and S. S. Li and Shuang Zhou and Tianyu Sun and X. Q. Li and Xiangyue Jin and Xiaojin Shen and Xiaosha Chen and Xinnan Song and Xinyi Zhou and Y. X. Zhu and Yanping Huang and Yaohui Li and Yi Zheng and Yuchen Zhu and Yunxian Ma and Zhen Huang and Zhipeng Xu and Zhongyu Zhang and Dongjie Ji and Jian Liang and Jianzhong Guo and Jin Chen and Leyi Xia and Miaojun Wang and Mingming Li and Peng Zhang and Ruyi Chen and Shangmian Sun and Shaoqing Wu and Shengfeng Ye and T. Wang and W. L. Xiao and Wei An and Xianzu Wang and Xiaowen Sun and Xiaoxiang Wang and Ying Tang and Yukun Zha and Zekai Zhang and Zhe Ju and Zhen Zhang and Zihua Qu},
      journal={CoRR},
      year={2025},
      volume={abs/2512.02556}, 
}

@misc{MiniMax-M2,
title = {MiniMax M2 and Agent: Ingenious in Simplicity},
author = {MiniMax},
year = {2025},
url = {https://www.minimax.io/news/minimax-m2}
}

@article{open_domain1,
  author       = {Adithya Bhaskar and
                  Xi Ye and
                  Danqi Chen},
  title        = {Language Models that Think, Chat Better},
  journal      = {CoRR},
  volume       = {abs/2509.20357},
  year         = {2025}
}

@article{open_domain2,
  author       = {Yuyuan Zeng and
                  Yufei Huang and
                  Can Xu and
                  Qingfeng Sun and
                  Jianfeng Yan and
                  Guanghui Xu and
                  Tao Yang and
                  Fengzong Lian},
  title        = {Zero Reinforcement Learning Towards General Domains},
  journal      = {CoRR},
  volume       = {abs/2510.25528},
  year         = {2025}
}

@inproceedings{cant_self_verify1,
  author       = {Kaya Stechly and
                  Karthik Valmeekam and
                  Subbarao Kambhampati},
  title        = {On the self-verification limitations of large language models on reasoning
                  and planning tasks},
  booktitle    = {{ICLR}},
  publisher    = {OpenReview.net},
  year         = {2025}
}

@inproceedings{cant_self_verify2,
  author       = {Ruixin Hong and
                  Hongming Zhang and
                  Xinyu Pang and
                  Dong Yu and
                  Changshui Zhang},
  title        = {A Closer Look at the Self-Verification Abilities of Large Language
                  Models in Logical Reasoning},
  booktitle    = {{NAACL-HLT}},
  pages        = {900--925},
  publisher    = {Association for Computational Linguistics},
  year         = {2024}
}

@inproceedings{cant_self_verify3,
  author       = {Yunxiang Zhang and
                  Muhammad Khalifa and
                  Lajanugen Logeswaran and
                  Jaekyeom Kim and
                  Moontae Lee and
                  Honglak Lee and
                  Lu Wang},
  title        = {Small Language Models Need Strong Verifiers to Self-Correct Reasoning},
  booktitle    = {{ACL} (Findings)},
  pages        = {15637--15653},
  publisher    = {Association for Computational Linguistics},
  year         = {2024}
}

@article{aha_1,
  author       = {Weihao Zeng and
                  Yuzhen Huang and
                  Qian Liu and
                  Wei Liu and
                  Keqing He and
                  Zejun Ma and
                  Junxian He},
  title        = {SimpleRL-Zoo: Investigating and Taming Zero Reinforcement Learning
                  for Open Base Models in the Wild},
  journal      = {CoRR},
  volume       = {abs/2503.18892},
  year         = {2025}
}

@article{aha_2,
  author       = {Jingcheng Hu and
                  Yinmin Zhang and
                  Qi Han and
                  Daxin Jiang and
                  Xiangyu Zhang and
                  Heung{-}Yeung Shum},
  title        = {Open-Reasoner-Zero: An Open Source Approach to Scaling Up Reinforcement
                  Learning on the Base Model},
  journal      = {CoRR},
  volume       = {abs/2503.24290},
  year         = {2025}
}

@article{fake_verification1,
  author       = {Jiachen Zhao and
                  Yiyou Sun and
                  Weiyan Shi and
                  Dawn Song},
  title        = {Can Aha Moments Be Fake? Identifying True and Decorative Thinking
                  Steps in Chain-of-Thought},
  journal      = {CoRR},
  volume       = {abs/2510.24941},
  year         = {2025}
}

@article{fake_verification2,
  author       = {Evelyn Yee and
                  Alice Li and
                  Chenyu Tang and
                  Yeon Ho Jung and
                  Ramamohan Paturi and
                  Leon Bergen},
  title        = {Dissociation of Faithful and Unfaithful Reasoning in LLMs},
  journal      = {CoRR},
  volume       = {abs/2405.15092},
  year         = {2024}
}

@article{cant_scale,
  author       = {Jack Lu and
                  Ryan Teehan and
                  Jinran Jin and
                  Mengye Ren},
  title        = {When Does Verification Pay Off? {A} Closer Look at LLMs as Solution
                  Verifiers},
  journal      = {CoRR},
  volume       = {abs/2512.02304},
  year         = {2025}
}

@article{join_train1,
  author       = {Xiaoyuan Liu and
                  Tian Liang and
                  Zhiwei He and
                  Jiahao Xu and
                  Wenxuan Wang and
                  Pinjia He and
                  Zhaopeng Tu and
                  Haitao Mi and
                  Dong Yu},
  title        = {Trust, But Verify: {A} Self-Verification Approach to Reinforcement
                  Learning with Verifiable Rewards},
  journal      = {CoRR},
  volume       = {abs/2505.13445},
  year         = {2025}
}

@article{join_train2,
  author       = {Fuxiang Zhang and
                  Jiacheng Xu and
                  Chaojie Wang and
                  Ce Cui and
                  Yang Liu and
                  Bo An},
  title        = {Incentivizing LLMs to Self-Verify Their Answers},
  journal      = {CoRR},
  volume       = {abs/2506.01369},
  year         = {2025}
}

@article{join_train3,
  author       = {Xiaoxuan Wang and
                  Bo Liu and
                  Song Jiang and
                  Jingzhou Liu and
                  Jingyuan Qi and
                  Xia Chen and
                  Baosheng He},
  title        = {From Solving to Verifying: {A} Unified Objective for Robust Reasoning
                  in LLMs},
  journal      = {CoRR},
  volume       = {abs/2511.15137},
  year         = {2025}
}

@inproceedings{GPT3,
  author       = {Tom B. Brown and
                  Benjamin Mann and
                  Nick Ryder and
                  Melanie Subbiah and
                  Jared Kaplan and
                  Prafulla Dhariwal and
                  Arvind Neelakantan and
                  Pranav Shyam and
                  Girish Sastry and
                  Amanda Askell and
                  Sandhini Agarwal and
                  Ariel Herbert{-}Voss and
                  Gretchen Krueger and
                  Tom Henighan and
                  Rewon Child and
                  Aditya Ramesh and
                  Daniel M. Ziegler and
                  Jeffrey Wu and
                  Clemens Winter and
                  Christopher Hesse and
                  Mark Chen and
                  Eric Sigler and
                  Mateusz Litwin and
                  Scott Gray and
                  Benjamin Chess and
                  Jack Clark and
                  Christopher Berner and
                  Sam McCandlish and
                  Alec Radford and
                  Ilya Sutskever and
                  Dario Amodei},
  title        = {Language Models are Few-Shot Learners},
  booktitle    = {NeurIPS},
  year         = {2020}
}

@article{GPT4,
  author       = {OpenAI},
  title        = {{GPT-4} Technical Report},
  journal      = {CoRR},
  volume       = {abs/2303.08774},
  year         = {2023}
}

@inproceedings{InstructGPT,
  author       = {Long Ouyang and
                  Jeffrey Wu and
                  Xu Jiang and
                  Diogo Almeida and
                  Carroll L. Wainwright and
                  Pamela Mishkin and
                  Chong Zhang and
                  Sandhini Agarwal and
                  Katarina Slama and
                  Alex Ray and
                  John Schulman and
                  Jacob Hilton and
                  Fraser Kelton and
                  Luke Miller and
                  Maddie Simens and
                  Amanda Askell and
                  Peter Welinder and
                  Paul F. Christiano and
                  Jan Leike and
                  Ryan Lowe},
  title        = {Training language models to follow instructions with human feedback},
  booktitle    = {NeurIPS},
  year         = {2022}
}

@article{llama1,
  author       = {Hugo Touvron and
                  Thibaut Lavril and
                  Gautier Izacard and
                  Xavier Martinet and
                  Marie{-}Anne Lachaux and
                  Timoth{\'{e}}e Lacroix and
                  Baptiste Rozi{\`{e}}re and
                  Naman Goyal and
                  Eric Hambro and
                  Faisal Azhar and
                  Aur{\'{e}}lien Rodriguez and
                  Armand Joulin and
                  Edouard Grave and
                  Guillaume Lample},
  title        = {LLaMA: Open and Efficient Foundation Language Models},
  journal      = {CoRR},
  volume       = {abs/2302.13971},
  year         = {2023}
}

@article{Qwen2.5,
  author       = {An Yang and
                  Baosong Yang and
                  Beichen Zhang and
                  Binyuan Hui and
                  Bo Zheng and
                  Bowen Yu and
                  Chengyuan Li and
                  Dayiheng Liu and
                  Fei Huang and
                  Haoran Wei and
                  Huan Lin and
                  Jian Yang and
                  Jianhong Tu and
                  Jianwei Zhang and
                  Jianxin Yang and
                  Jiaxi Yang and
                  Jingren Zhou and
                  Junyang Lin and
                  Kai Dang and
                  Keming Lu and
                  Keqin Bao and
                  Kexin Yang and
                  Le Yu and
                  Mei Li and
                  Mingfeng Xue and
                  Pei Zhang and
                  Qin Zhu and
                  Rui Men and
                  Runji Lin and
                  Tianhao Li and
                  Tingyu Xia and
                  Xingzhang Ren and
                  Xuancheng Ren and
                  Yang Fan and
                  Yang Su and
                  Yichang Zhang and
                  Yu Wan and
                  Yuqiong Liu and
                  Zeyu Cui and
                  Zhenru Zhang and
                  Zihan Qiu},
  title        = {Qwen2.5 Technical Report},
  journal      = {CoRR},
  volume       = {abs/2412.15115},
  year         = {2024}
}

@article{GRPO,
  author       = {Zhihong Shao and
                  Peiyi Wang and
                  Qihao Zhu and
                  Runxin Xu and
                  Junxiao Song and
                  Mingchuan Zhang and
                  Y. K. Li and
                  Y. Wu and
                  Daya Guo},
  title        = {DeepSeekMath: Pushing the Limits of Mathematical Reasoning in Open
                  Language Models},
  journal      = {CoRR},
  volume       = {abs/2402.03300},
  year         = {2024}
}

@article{search-r1,
  author       = {Bowen Jin and
                  Hansi Zeng and
                  Zhenrui Yue and
                  Dong Wang and
                  Hamed Zamani and
                  Jiawei Han},
  title        = {Search-R1: Training LLMs to Reason and Leverage Search Engines with
                  Reinforcement Learning},
  journal      = {CoRR},
  volume       = {abs/2503.09516},
  year         = {2025}
}

@article{ragen,
  author       = {Zihan Wang and
                  Kangrui Wang and
                  Qineng Wang and
                  Pingyue Zhang and
                  Linjie Li and
                  Zhengyuan Yang and
                  Xing Jin and
                  Kefan Yu and
                  Minh Nhat Nguyen and
                  Licheng Liu and
                  Eli Gottlieb and
                  Yiping Lu and
                  Kyunghyun Cho and
                  Jiajun Wu and
                  Li Fei{-}Fei and
                  Lijuan Wang and
                  Yejin Choi and
                  Manling Li},
  title        = {{RAGEN:} Understanding Self-Evolution in {LLM} Agents via Multi-Turn
                  Reinforcement Learning},
  journal      = {CoRR},
  volume       = {abs/2504.20073},
  year         = {2025}
}

@article{diverse1,
  author       = {Yi Su and
                  Dian Yu and
                  Linfeng Song and
                  Juntao Li and
                  Haitao Mi and
                  Zhaopeng Tu and
                  Min Zhang and
                  Dong Yu},
  title        = {Crossing the Reward Bridge: Expanding {RL} with Verifiable Rewards
                  Across Diverse Domains},
  journal      = {CoRR},
  volume       = {abs/2503.23829},
  year         = {2025}
}

@article{diverse2,
  author       = {Tianyu Yu and
                  Bo Ji and
                  Shouli Wang and
                  Shu Yao and
                  Zefan Wang and
                  Ganqu Cui and
                  Lifan Yuan and
                  Ning Ding and
                  Yuan Yao and
                  Zhiyuan Liu and
                  Maosong Sun and
                  Tat{-}Seng Chua},
  title        = {{RLPR:} Extrapolating {RLVR} to General Domains without Verifiers},
  journal      = {CoRR},
  volume       = {abs/2506.18254},
  year         = {2025}
}

@article{diverse3,
  author       = {Anisha Gunjal and
                  Anthony Wang and
                  Elaine Lau and
                  Vaskar Nath and
                  Bing Liu and
                  Sean Hendryx},
  title        = {Rubrics as Rewards: Reinforcement Learning Beyond Verifiable Domains},
  journal      = {CoRR},
  volume       = {abs/2507.17746},
  year         = {2025}
}

@article{efficient1,
  author       = {Yang Sui and
                  Yu{-}Neng Chuang and
                  Guanchu Wang and
                  Jiamu Zhang and
                  Tianyi Zhang and
                  Jiayi Yuan and
                  Hongyi Liu and
                  Andrew Wen and
                  Shaochen Zhong and
                  Na Zou and
                  Hanjie Chen and
                  Xia Hu},
  title        = {Stop Overthinking: {A} Survey on Efficient Reasoning for Large Language
                  Models},
  journal      = {Trans. Mach. Learn. Res.},
  volume       = {2025},
  year         = {2025}
}

@article{efficient2,
  author       = {Sicheng Feng and
                  Gongfan Fang and
                  Xinyin Ma and
                  Xinchao Wang},
  title        = {Efficient Reasoning Models: {A} Survey},
  journal      = {Trans. Mach. Learn. Res.},
  volume       = {2025},
  year         = {2025}
}

@article{efficient3,
  author       = {Rui Wang and
                  Hongru Wang and
                  Boyang Xue and
                  Jianhui Pang and
                  Shudong Liu and
                  Yi Chen and
                  Jiahao Qiu and
                  Derek Fai Wong and
                  Heng Ji and
                  Kam{-}Fai Wong},
  title        = {Harnessing the Reasoning Economy: {A} Survey of Efficient Reasoning
                  for Large Language Models},
  journal      = {CoRR},
  volume       = {abs/2503.24377},
  year         = {2025}
}

@article{stable1,
  author       = {Zhicheng Yang and
                  Zhijiang Guo and
                  Yinya Huang and
                  Yongxin Wang and
                  Dongchun Xie and
                  Yiwei Wang and
                  Xiaodan Liang and
                  Jing Tang},
  title        = {Depth-Breadth Synergy in {RLVR:} Unlocking {LLM} Reasoning Gains with
                  Adaptive Exploration},
  journal      = {CoRR},
  volume       = {abs/2508.13755},
  year         = {2025}
}

@article{effective1,
  author       = {Fang Wu and
                  Weihao Xuan and
                  Ximing Lu and
                  Za{\"{\i}}d Harchaoui and
                  Yejin Choi},
  title        = {The Invisible Leash: Why {RLVR} May Not Escape Its Origin},
  journal      = {CoRR},
  volume       = {abs/2507.14843},
  year         = {2025}
}

@article{effective2,
  author       = {Zhipeng Chen and
                  Xiaobo Qin and
                  Youbin Wu and
                  Yue Ling and
                  Qinghao Ye and
                  Wayne Xin Zhao and
                  Guang Shi},
  title        = {Pass@k Training for Adaptively Balancing Exploration and Exploitation
                  of Large Reasoning Models},
  journal      = {CoRR},
  volume       = {abs/2508.10751},
  year         = {2025}
}

@article{verithinker,
  author       = {Zigeng Chen and
                  Xinyin Ma and
                  Gongfan Fang and
                  Ruonan Yu and
                  Xinchao Wang},
  title        = {VeriThinker: Learning to Verify Makes Reasoning Model Efficient},
  journal      = {CoRR},
  volume       = {abs/2505.17941},
  year         = {2025}
}

@article{test_time_scaling,
  author       = {Charlie Snell and
                  Jaehoon Lee and
                  Kelvin Xu and
                  Aviral Kumar},
  title        = {Scaling {LLM} Test-Time Compute Optimally can be More Effective than
                  Scaling Model Parameters},
  journal      = {CoRR},
  volume       = {abs/2408.03314},
  year         = {2024}
}

@article{reward_model_survey,
  author       = {Jialun Zhong and
                  Wei Shen and
                  Yanzeng Li and
                  Songyang Gao and
                  Hua Lu and
                  Yicheng Chen and
                  Yang Zhang and
                  Wei Zhou and
                  Jinjie Gu and
                  Lei Zou},
  title        = {A Comprehensive Survey of Reward Models: Taxonomy, Applications, Challenges,
                  and Future},
  journal      = {CoRR},
  volume       = {abs/2504.12328},
  year         = {2025}
}

@inproceedings{reward_model_survey2,
  author       = {Rui Yu and
                  Shenghua Wan and
                  Yucen Wang and
                  Chen{-}Xiao Gao and
                  Le Gan and
                  Zongzhang Zhang and
                  De{-}Chuan Zhan},
  title        = {Reward Models in Deep Reinforcement Learning: {A} Survey},
  booktitle    = {{IJCAI}},
  pages        = {10807--10816},
  publisher    = {ijcai.org},
  year         = {2025}
}

@article{scalar_reward1,
  author       = {Chris Yuhao Liu and
                  Liang Zeng and
                  Yuzhen Xiao and
                  Jujie He and
                  Jiacai Liu and
                  Chaojie Wang and
                  Rui Yan and
                  Wei Shen and
                  Fuxiang Zhang and
                  Jiacheng Xu and
                  Yang Liu and
                  Yahui Zhou},
  title        = {Skywork-Reward-V2: Scaling Preference Data Curation via Human-AI Synergy},
  journal      = {CoRR},
  volume       = {abs/2507.01352},
  year         = {2025}
}

@article{scalar_reward2,
  author       = {Chris Yuhao Liu and
                  Liang Zeng and
                  Jiacai Liu and
                  Rui Yan and
                  Jujie He and
                  Chaojie Wang and
                  Shuicheng Yan and
                  Yang Liu and
                  Yahui Zhou},
  title        = {Skywork-Reward: Bag of Tricks for Reward Modeling in LLMs},
  journal      = {CoRR},
  volume       = {abs/2410.18451},
  year         = {2024}
}

@inproceedings{GRM1,
  author       = {Lunjun Zhang and
                  Arian Hosseini and
                  Hritik Bansal and
                  Mehran Kazemi and
                  Aviral Kumar and
                  Rishabh Agarwal},
  title        = {Generative Verifiers: Reward Modeling as Next-Token Prediction},
  booktitle    = {{ICLR}},
  publisher    = {OpenReview.net},
  year         = {2025}
}

@article{GRM2,
  author       = {Dakota Mahan and
                  Duy Phung and
                  Rafael Rafailov and
                  Chase Blagden and
                  Nathan Lile and
                  Louis Castricato and
                  Jan{-}Philipp Fr{\"{a}}nken and
                  Chelsea Finn and
                  Alon Albalak},
  title        = {Generative Reward Models},
  journal      = {CoRR},
  volume       = {abs/2410.12832},
  year         = {2024}
}

@article{GRM3,
  author       = {Zijun Liu and
                  Peiyi Wang and
                  Runxin Xu and
                  Shirong Ma and
                  Chong Ruan and
                  Peng Li and
                  Yang Liu and
                  Yu Wu},
  title        = {Inference-Time Scaling for Generalist Reward Modeling},
  journal      = {CoRR},
  volume       = {abs/2504.02495},
  year         = {2025}
}

@article{rm-r1,
  author       = {Xiusi Chen and
                  Gaotang Li and
                  Ziqi Wang and
                  Bowen Jin and
                  Cheng Qian and
                  Yu Wang and
                  Hongru Wang and
                  Yu Zhang and
                  Denghui Zhang and
                  Tong Zhang and
                  Hanghang Tong and
                  Heng Ji},
  title        = {{RM-R1:} Reward Modeling as Reasoning},
  journal      = {CoRR},
  volume       = {abs/2505.02387},
  year         = {2025}
}

@article{rewardanything,
  author       = {Zhuohao Yu and
                  Jiali Zeng and
                  Weizheng Gu and
                  Yidong Wang and
                  Jindong Wang and
                  Fandong Meng and
                  Jie Zhou and
                  Yue Zhang and
                  Shikun Zhang and
                  Wei Ye},
  title        = {RewardAnything: Generalizable Principle-Following Reward Models},
  journal      = {CoRR},
  volume       = {abs/2506.03637},
  year         = {2025}
}

@inproceedings{rm-dpo,
  author       = {Changyu Chen and
                  Zichen Liu and
                  Chao Du and
                  Tianyu Pang and
                  Qian Liu and
                  Arunesh Sinha and
                  Pradeep Varakantham and
                  Min Lin},
  title        = {Bootstrapping Language Models with {DPO} Implicit Rewards},
  booktitle    = {{ICLR}},
  publisher    = {OpenReview.net},
  year         = {2025}
}

@article{PPO,
  author       = {John Schulman and
                  Filip Wolski and
                  Prafulla Dhariwal and
                  Alec Radford and
                  Oleg Klimov},
  title        = {Proximal Policy Optimization Algorithms},
  journal      = {CoRR},
  volume       = {abs/1707.06347},
  year         = {2017}
}

@article{DAPO,
  author       = {Qiying Yu and
                  Zheng Zhang and
                  Ruofei Zhu and
                  Yufeng Yuan and
                  Xiaochen Zuo and
                  Yu Yue and
                  Tiantian Fan and
                  Gaohong Liu and
                  Lingjun Liu and
                  Xin Liu and
                  Haibin Lin and
                  Zhiqi Lin and
                  Bole Ma and
                  Guangming Sheng and
                  Yuxuan Tong and
                  Chi Zhang and
                  Mofan Zhang and
                  Wang Zhang and
                  Hang Zhu and
                  Jinhua Zhu and
                  Jiaze Chen and
                  Jiangjie Chen and
                  Chengyi Wang and
                  Hongli Yu and
                  Weinan Dai and
                  Yuxuan Song and
                  Xiangpeng Wei and
                  Hao Zhou and
                  Jingjing Liu and
                  Wei{-}Ying Ma and
                  Ya{-}Qin Zhang and
                  Lin Yan and
                  Mu Qiao and
                  Yonghui Wu and
                  Mingxuan Wang},
  title        = {{DAPO:} An Open-Source {LLM} Reinforcement Learning System at Scale},
  journal      = {CoRR},
  volume       = {abs/2503.14476},
  year         = {2025}
}

@misc{aime24,
      title={American Invitational Mathematics Examination (AIME) 2024}, 
      author={Zhang, Yifan and Math-AI, Team},
      year={2024},
}

@misc{aime25,
      title={American Invitational Mathematics Examination (AIME) 2025}, 
      author={Zhang, Yifan and Math-AI, Team},
      year={2025},
}

@inproceedings{OlympiaBench,
  author       = {Chaoqun He and
                  Renjie Luo and
                  Yuzhuo Bai and
                  Shengding Hu and
                  Zhen Leng Thai and
                  Junhao Shen and
                  Jinyi Hu and
                  Xu Han and
                  Yujie Huang and
                  Yuxiang Zhang and
                  Jie Liu and
                  Lei Qi and
                  Zhiyuan Liu and
                  Maosong Sun},
  title        = {OlympiadBench: {A} Challenging Benchmark for Promoting {AGI} with
                  Olympiad-Level Bilingual Multimodal Scientific Problems},
  booktitle    = {{ACL} {(1)}},
  pages        = {3828--3850},
  publisher    = {Association for Computational Linguistics},
  year         = {2024}
}

@inproceedings{Math500,
  author       = {Hunter Lightman and
                  Vineet Kosaraju and
                  Yuri Burda and
                  Harrison Edwards and
                  Bowen Baker and
                  Teddy Lee and
                  Jan Leike and
                  John Schulman and
                  Ilya Sutskever and
                  Karl Cobbe},
  title        = {Let's Verify Step by Step},
  booktitle    = {{ICLR}},
  publisher    = {OpenReview.net},
  year         = {2024}
}

@inproceedings{verl,
  author       = {Guangming Sheng and
                  Chi Zhang and
                  Zilingfeng Ye and
                  Xibin Wu and
                  Wang Zhang and
                  Ru Zhang and
                  Yanghua Peng and
                  Haibin Lin and
                  Chuan Wu},
  title        = {HybridFlow: {A} Flexible and Efficient {RLHF} Framework},
  booktitle    = {EuroSys},
  pages        = {1279--1297},
  publisher    = {{ACM}},
  year         = {2025}
}
\bibliographystyle{icml2025}

\newpage
\appendix
\onecolumn


\end{document}